# An IoT Framework for Heart Disease Prediction based on MDCNN Classifier

**Mohammad Ayoub Khan, Senior Member, IEEE**
College of Computing and Information Technology, University of Bisha, Bisha, 67714, Kingdom of Saudi Arabia

Corresponding author: Mohammad Ayoub Khan (e-mail: ayoub.khan@ieee.org).

**ABSTRACT** Nowadays, heart disease is the leading cause of death worldwide. Predicting heart disease is a complex task since it requires experience along with advanced knowledge. Internet of Things (IoT) technology has lately been adopted in healthcare systems to collect sensor values for heart disease diagnosis and prediction. Many researchers have focused on the diagnosis of heart disease, yet the accuracy of the diagnosis results is low. To address this issue, an IoT framework is proposed to evaluate heart disease more accurately using a Modified Deep Convolutional Neural Network (MDCNN). The smartwatch and heart monitor device that is attached to the patient monitors the blood pressure and electrocardiogram (ECG). The MDCNN is utilized for classifying the received sensor data into normal and abnormal. The performance of the system is analyzed by comparing the proposed MDCNN with existing deep learning neural networks and logistic regression. The results demonstrate that the proposed MDCNN based heart disease prediction system performs better than other methods. The proposed method shows that for the maximum number of records, the MDCNN achieves an accuracy of 98.2 which is better than existing classifiers.

## I. INTRODUCTION

The internet of things (IoT) along with wearable monitoring systems is a rising technology that is anticipated to contribute an extensive range of healthcare applications [1, 27, 28, 32,33, 37]. The healthcare industry was fast to adopt the IoT [2,3] as integrating IoT aspects into medical devices enhances the quality as well as the efficiency of service. This brings remarkable advantages for older people, patients with chronic conditions, and individuals needing stable management [4]. IoT-based healthcare applications are utilized to gather essential data, such as real-time changes in health parameters and updates of medical parameter's severity within a standard time interval, so IoT devices incessantly produce huge amounts of health data. The IoT is acknowledged as the most crucial future technology and is gaining much attention from healthcare industries [5, 38, 39]. The IoT was revealed to have important potential in high-risk environment, health, and safety (EHS) [6] industries, where people's lives are at risk and IoT-based applications are primed to present safe, dependable, and effective solutions [7]. Furthermore, IoT-based remote health monitoring systems nurse the health of elderly people who do not want to compromise their ease and who prefer to stay at home [8].

The incorporation of the IoT in the healthcare industry has directed researchers around the globe to develop smart applications, for instance, mobile healthcare, health-aware suggestions, and intelligent healthcare systems [9, 40]. The size of internet of things healthcare market is estimated approximately of net worth $136 billion by 2021[41]. The market is expected to elevate with a compound annual growth rate (CAGR) of 23.4% by 2024[41]. Smart wearable gadgets can be utilized for patients who need to gather data about their health condition, for instance, heart rate, blood pressure, and glucose level. These data can be constantly monitored via sensors on the wearable equipment and sent to smartphones [10]. Electrocardiogram (ECG) [11] sensor nodes are joined to the IoT network that is backed up as a result of the plug-and-play functionality [12]. Via IoT systems, the amassed data is stored in the cloud server [13]. Consequently, real-

time data along with historical data can be accessed distantly [14]. Healthcare scrutinizing ought to be incessant to track the patient's body parameters as well as offer steady and dependable data to the doctors or the medical team for diagnosis [15]. The vital parameters registered by healthcare monitoring are heart rate, temperature, blood pressure, weight, glucose level, and ECG. Personal healthcare utilizing IoT devices will offer a means to a healthy life with a lower price [16].

Heart disease is a grave disease that influences the heart's functionality and gives rise to complications such as infection of the coronary artery and diminished blood vessel function [17, 34].

Heart disease patients do not feel sick until the very last stage of the disease, and then it is too late because the damages have become irretrievable [18, 26, 35]. In the past years, sensor networks for healthcare IoT have advanced quickly, so it is now possible to incorporate instantaneous health data by linking bodies and sensors [19]. It is important to diagnose patients early by means of ECG. IoT-based heart attack detection systems [20, 29, 30, 31] raise privacy and security concerns. As mobile devices are possible targets for malevolent attacks, more studies are needed on safety countermeasures. A fault-tolerant algorithm should be developed for a dependable IoT system [21].

Long short-term memory (LSTM) is an uncommon sort of intermittent neural system, which may have the option to associate past data to the present task [45, 46]. From those temporal data the improvement of disease in the patient can be predicted. Whereas the convolution neural network is used to design the system for prediction of disease. The accuracy of model will be decreased if there is any missing data. To overcome this issue modified deep convolutional neural network is proposed.

The contribution of this work is as follow: Firstly, contribution is towards framework design for architecture of heart disease prediction system. The framework proposes the modules in predictions system such as algorithms, long-range (LoRa) communication protocols, LoRa cloud, servers and dataset. Secondly, we proposed modified deep convolutional neural network. The convolution neural network is used to design the system for prediction of heart disease. The performance of the proposed algorithms has been evaluated and compared to existing work.

The structure of this paper is as follows. Section II presents studies related to the proposed technique. Section III briefly discusses the proposed methodology, and Section IV explores the experiment's outcome and analysis. Section V provides conclusions.

## II. RELATED WORK

Abdel-Basset et al. [16] presented a framework based on IoT and computer-supported diagnoses to observe persons with heart failure, with the data taken from various sources. Initially, the data from the body sensors about heart failure symptoms was taken by the users' mobiles via Bluetooth technology and transmitted by a smart gateway to the cloud database. Clinicians categorized the patients into multiple groups depending on their symptoms. Finally, the IoT and neutrosophic multi criteria decision making (NMCDM) technique was incorporated to recognize, observe, and control heart failures with minimal cost and time to evaluate the disease. The experiential outcomes corroborated the performance of the high-level system.

Kumar and Gandhi [17] recommended a scalable three-tier architecture for processing and storing copious wearable sensor data. Tier 1 took care of the compilation of the data from the IoT wearable sensor devices. Tier 2 utilized Apache HBase for effectively storing the wearable IoT sensor data in a cloud computing environment. Subsequently, Tier 3 utilized Apache Mahout to build the logistic-regression-based prediction framework for the heart disease. Finally, a receiver operating characteristic (ROC) analysis was carried out to recognize the noteworthy clinical parameters of heart disease.

Kumar et al. [18] recommended a cloud and IoT-based mobile healthcare application for observing and diagnosing grave diseases. The system comprises eight major units: medical IoT devices, medical records, a University of California Irvine (UCI) repository dataset, a data collection module, a cloud database, a secured storage mechanism, a knowledge base, and a health prediction and diagnosing structure. This framework utilized a fuzzy temporal neural classifier for estimating health conditions. The experiential outcomes showed that the system outperforms other frameworks.

Ali et al. [19] introduced an automated diagnostic framework for heart disease diagnosis. First, normalization was done on feature vectors, and then the data were divided into training and test datasets. Subsequently, selection and feature ranking were done on the trained data using a statistical framework. The same subset of features that was chosen by the framework in the training phase was utilized for testing the data. The training data with a reduced number of features were employed by a neural network (NN) for training purposes. The performance of the trained NN was assessed using the test data.

Gupta et al. [20] put forward an IoT-based cloud architecture. The presented system utilized the embedded sensors of the equipment instead of smartphone sensors or wearable sensors for saving the values of the basic health-associated parameters. The cloud-based architecture comprises the private cloud, the cloud data center (CDC), and the public cloud. This architecture utilized XML Web services for the fast and secure communication of data. It could be perceived that the overall response between the CDC and the local database server remains almost consistent with the rise in the number of users.

Rathee et al. [21] proposed a secure healthcare framework based on blockchain methodology. The blockchain was utilized for assuring the transparency and security of document accessibility, patient records, and the shipment process among providers and customers. The experiential analysis of the framework was gauged on the illegal actions or communications by malicious IoT objects.

Vijayashree and Sultana [22] introduced a function for recognizing the optimum weights based on population diversity and tuning functions. In addition, the presented framework built a fitness function for particle swarm optimizations (PSOs) with support vector machines (SVMs). The PSO-SVM feature selection algorithm recognized six notable features for categorizing heart disease: sex, maximum heart rate, fasting blood sugar level, resting ECG, multiple major vessels, and exercise-induced angina. The performance of the suggested PSO-SVM system was compared with that of several existing methods. The suggested methodology was shown to perform better than other methods.

Mutlag et al. [23] identified the issues, challenges, and difficulties involved in healthcare IoT systems and offered various suggestions to resolve potential and current resource management difficulties by adopting three main factors: load balancing, interoperability, and computation offloading.

X Liu et al. [43] proposed system contains two subsystems: the RFRS feature selection system and a classification system with an ensemble classifier. The first system includes three stages: (i) data discretization, (ii) feature extraction using the Relief FS algorithm, and (iii) feature reduction using the heuristic rough set reduction algorithm that we developed. In the second system, an ensemble classifier is proposed based on the C4.5 classifier. The Statlog (Heart) dataset, obtained from the UCI database, was used for experiments.

Haq et al. [44] have developed a machine-learning-based diagnosis system for heart disease prediction by using heart disease dataset. They used seven popular machine learning algorithms, three feature selection algorithms, the cross-validation method, and seven classifiers performance evaluation metrics such as classification accuracy, specificity, sensitivity, Matthews' correlation coefficient, and execution time. The proposed system can easily identify and classify people with heart disease from healthy people.

Senthilkumar et al. [53] proposed a method for predicting cardiovascular disease by hybrid machine learning. Authors have coined hybrid random forest with liner model (HRFLM) that have achieved 88.7 of accuracy.

The conventional Neural Network is an advanced version of Neural Network and it has serval layers to learn the lower and higher-level features. In 2006, Hinton et al. [47]. built up another algorithm to train the neuron layers of profound design, which they called layer wise preparing. This learning calculation is viewed as a solo single layer avariciously preparing where a deep network is trained layer by layer. Since this strategy turned out to be progressively compelling, it has been begun to be utilized for preparing numerous deep network systems. One of the most dominant profound systems is the convolutional neural system that can incorporate numerous shrouded layers performing convolution and subsampling so as to extricate low to significant levels of highlights of the features. This system has indicated an incredible productivity in various territories, especially, in computer vision, organic calculation, unique mark improvement, etc. Fundamentally, this kind of systems comprises of three layers: convolution layers, subsampling or pooling layers, and full association layers. LSTM adapts long term conditions by fusing a memory cell that can save state after some time. Three gates are prepared in LSTM for choosing which data to outline or overlook before proceeding onward to the following subsequence. LSTM is appropriate to catch successive data from transient information and has demonstrated preferences in machine interpretation, discourse acknowledgment, and picture subtitling, and so forth. In the restorative space, numerous endeavors have been made to apply LSTM for clinical expectation dependent on electronic wellbeing records. Zouka and Hosni [24] put forward a secure lightweight authentication structure that shields personal health data and assures secure communication. The suggested structure enables doctors to monitor patients' real-time biosignals and was fitted with an emergency rescue approach utilizing a machine-to-machine (M2M) patient monitoring screen and remote health app. The outcomes corroborated that the suggested structure obtained high-level results as it diminished the overhead of the access time. The structure has the highest key generation time when the verification time and transfer time are considered.

To address this issue, an IoT framework is proposed to evaluate heart disease more accurately using a modified deep convolutional neural network (MDCNN). The smartwatch and heart monitor device that is attached to the patient monitors the blood pressure, and ECG. The MDCNN is utilized for classifying the received sensor data into normal and abnormal. The performance of the system is analyzed by comparing the proposed MDCNN with existing deep learning neural networks and logistic regression

## II. PROPOSED METHODOLOGY

The classification is performed utilizing MDCNN based on the received sensor data for evaluating the heart disease. To do the classification, the system undertakes training as well as testing. The data from the UCI machine learning repository, Framingham, and Public Health Dataset were utilized for training and evaluating the disease [25, 48]. The UCI repository has Cleveland, Hungary, Switzerland, and the VA Long Beach databases. However, we have selected the Cleveland database as it has 303 records which are most complete. The pre-processing is performed in the training phase. Then, a feature selection mechanism is executed utilizing the mapping-based cuttlefish optimization algorithm. Subsequently, classification is performed on the selected feature data.

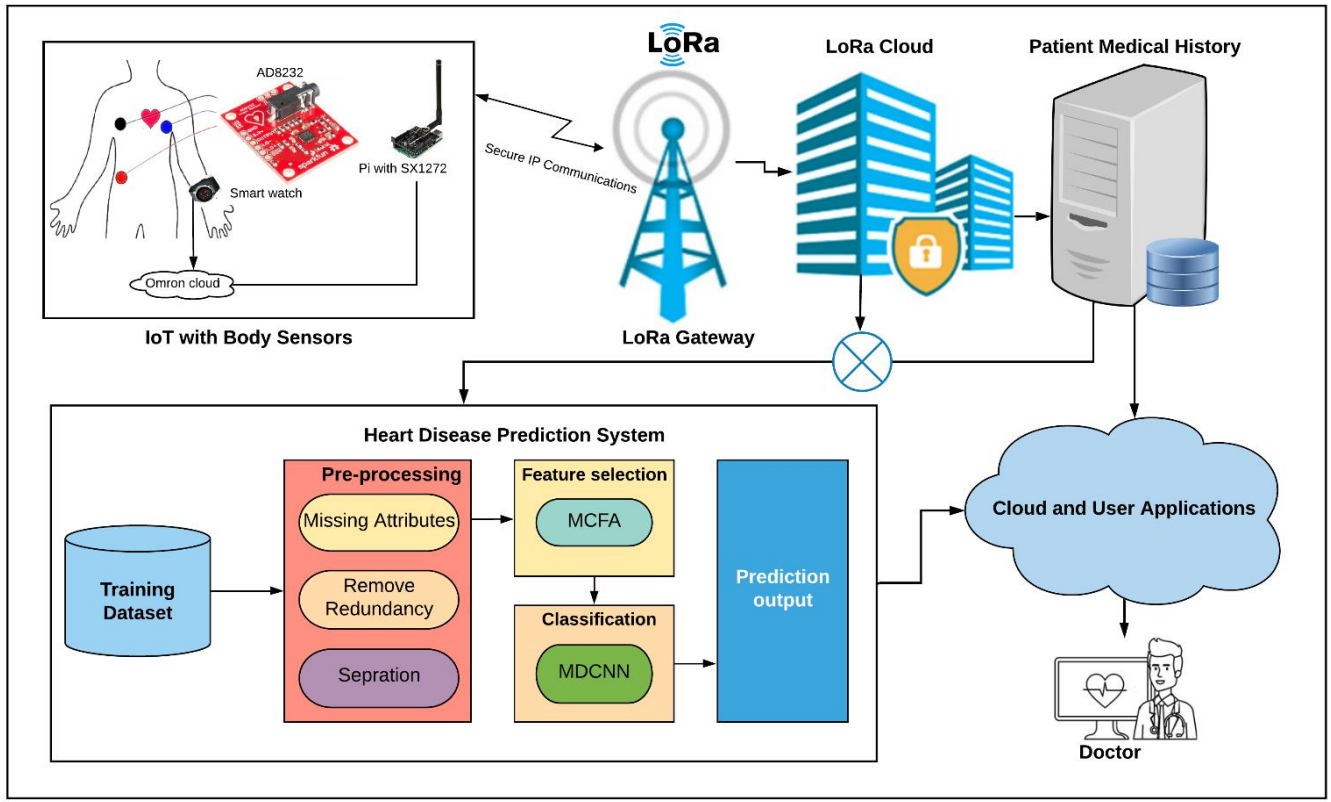

**FIGURE 1.** Proposed framework of heart disease prediction system

After training, the sensor information from the long range (LoRa) [42] cloud server is tested and classified as normal or abnormal. If the outcome is abnormal, an alert message is delivered to the doctor to treat the patient. The proposed framework is exhibited in Fig. 1.

In the proposed method, heart disease of the patient can be predicted utilizing a MDCNN. For this, the system undergoes training and testing. For training, the data from the UCI, Framingham, Public Health Dataset dataset is utilized, and the MDCNN classifier is employed to train the system. The UCI dataset encompasses the classified outcomes (normal and abnormal) of the data. If the sensor values from the IoT are tested directly, then it takes time to detect the disease, and there is a likelihood of obtaining an erroneous outcome. Thus, the system undergoes training. The proposed MDCNN classifier for training follows three processes: (i) pre-processing, (ii) feature selection, and (iii) classification. The outcome of the training encompasses two classes: (a) normal (the patient's heart condition is normal) and (b) abnormal (the patient's heart condition is abnormal). After training, the testing phase is executed. The sensor device joined to the patient continuously sends sensor values. These are classified based on the training outcomes, which means that the sensor values from the IoT are contrasted with the values of the training phase. The system compares the values and provides classified outcomes. Below, the steps in the training phase are explained.

### A. Pre-Processing

This is the first stage of the diagnosis process. It embraces three steps: replacement of missing attributes, removal of redundancy, and separation. The missing value of a specific attribute is replaced after checking the entire patient's age group, cholesterol, and blood pressure. If most of the attribute values of a patient match, then the value is substituted in the same position. The redundancy removal reduces the data size through the elimination of redundant (irrelevant) attributes. Next, the patients are separated based on the type of chest pain type they have: (1) typical angina, (2) atypical angina, (3) non-anginal pain, and (4) asymptomatic pain.

### B. Feature Selection

The significant features for evaluating heart disease are chosen utilizing the mapping-based cuttlefish optimization algorithm (MCFA) [36]. The population-based search called the cuttlefish algorithm (CFA) mimics the mechanisms of skin color changing in cuttlefish. The colors and patterns seen in cuttlefish are generated by light reflected from different layers of cells encompassing chromatophores, leucophores, and iridophores. The CFA uses two main processes, visibility and reflection, as the search strategy for finding the optimal solution. The reflection process simulates light reflections, and the visibility process simulates the visibility of matching patterns. An important feature of the optimization algorithm is its ability to integrate local and global search. This feature could adjust the contribution of local and global search in the initial step and during the search process. The main problem is that optimization algorithms sometimes get trapped into a local optimum. A solution is to apply chaos mapping, which brings instability and dynamism to the algorithm by enforcing a random search. This solution helps the algorithm to avert local minima. Therefore, an MCFA is proposed by adopting chaos mapping in the CFA, which ameliorates the algorithm's performance in handling different optimization problems.

First, in this method, a chaos-mapping-based population initialization is done. A search supported by chaotic mappings has the possibility of accessing most states in a certain zone without any iteration. The chaotic mapping that is chosen to produce chaos sequences in the proposed MCFA is written as in equation (1) and (2).

$$Cr_{n+1} = \delta * Cr_n * (1 - Cr_n) \text{ for } 0 < \delta \leq 4 \quad (1)$$

$$Br_{n+1} = \delta * Br_n * (1 - Br_n) \text{ for } 0 < \delta \leq 4 \quad (2)$$

where $\delta$ indicates the function's initial value, $Cr$ and $Br$ represent a function based on the activities of the logistic map varying from 0 to 1. Utilizing this framework improves the diversity of the solutions and their coverage. The $Cr$ and $Br$ parameters are employed instead of a random parameter used in the CFA. After chaotic initialization, the new solution is found utilizing visibility and reflection, as shown in equation (3).

$$N_s = R_e + V_s \quad (3)$$

where $N_s$ indicates the new solution, $R_e$ denotes the reflection, and $V_s$ denotes visibility. This algorithm splits the population into four categories. For Category 1, the algorithm applies Case 1 and Case 2 (interaction between iridophores and chromatophores) to produce a new solution. These two cases work as a global search utilizing the value of each point to determine a new area around the best solution with a particular interval. For Category 2, the algorithm utilizes Case 3 (reflection operator of iridophores) and Case 4 (interaction between iridophores and chromatophores), which work as a local search to produce a new solution. For Category 3, Case 5 (interaction between chromatophores and leucophores) is employed to create solutions nearer the best solution (local search). Finally, for Category 3, Case 6 (reflection operator of leucophores) is implemented as a global search by reflecting any incoming light without any modification. The equations that are utilized to evaluate the reflection and visibility for the four categories are described below.

a) **Cases 1 and 2 for Category 1**

$$R_e[j] = Cr * G_1[i].points[j] \quad (4)$$

$$V_s[j] = Br * (b.points[j] - G_1[i].points[j]) \quad (5)$$

where $G_1$ is Category 1, which presents an element in $G$; $i$ is the $i^{th}$ cell in $G_1$; $points[j]$ represents the $j^{th}$ point of the $i^{th}$ element in group $G$; and $b$ is the best solution in which the mean value of the best points is evaluated.

Further, $Cr$ and $Br$ are two chaos random numbers, $R_e$ denotes the degree of reflection, and $V_s$ indicates the degree of visibility of the final pattern.

b) **Cases 3 and 4 for Category 2**

$$R_e[j] = Cr * b.points[j] \quad (6)$$

$$V_s[j] = Br * (b.points[j] - G_2[i].points[j]) \quad (7)$$

c) **Cases 5 for Category 3**

$$R_e[j] = Cr * G_1[j].points[j] \quad (8)$$

$$V_s[j] = Br * (b.points[j] - AV_B) \quad (9)$$

where $AV_B$ specifies the average value of the best points.

d) **Cases 6 for Category 4**

$$p[i] \cdot Po\ int\ s\ [j] = random * (U_l - L_l) + L_l \quad (10)$$

where $i, j = 1,2,\ldots\ldots n$ and $U_l$ and $L_l$ indicate the upper and lower limits of the problem domain.

### C. Classification Using the MDCNN Classifier

In the deep learning neural network (DLNN), the weight values are optimized using the adaptive elephant herd optimization (AEHO) algorithm, so it is called as MDCNN. After feature selection, the chosen features are classified utilizing the MDCNN. For this, each selected feature is provided as input for the MDCNN classifier. The weights are arbitrarily assigned values and are linked with each input. The hidden nodes of the subsequent hidden layer perform the function of adding the product of the input value and the weight vector of all the input nodes that are linked to it. Random weight values improve the backpropagation process to acquire the result. In this way, the optimization is performed. The activation operation is then employed, and this layer's output is transported to the consecutive layer. These weights strongly influence the classifier's output. The algorithmic steps in the MDCNN classification are the following.

**Step 1:** Let the selected feature values and their equivalent weights be expressed using equations (11) and (12):

$$F_i = \{F_1, F_2, F_3, \ldots, F_n\} \quad (11)$$

$$W_i = \{W_1, W_2, W_3, \ldots, W_n\} \quad (12)$$

where $F_i$ indicates the input value, which denotes $n$ chosen features, $F_1, F_2, F_3, \ldots, F_n$, and $W_i$ denotes the weight value of $F_i$, which specifies the $n$ corresponding weights, $W_1, W_2, W_3, \ldots, W_n$.

**Step 2:** Multiply the inputs with the arbitrarily chosen weight vectors and then add them up:

$$M = \sum_{i-1}^{n} F_i W_i \quad (13)$$

where $M$ denotes the summed value.

**Step 3:** Determine the activation function (AF).

$$A_{f_i} = C_i(\sum_{i=1}^{n} F_i W_i) \quad (14)$$

$$C_i = e^{-F_i^2} \quad (15)$$

Here, $A_{f_i}$ specifies the activation function, whereas $C_i$ specifies the exponential of $F_i$. A Gaussian function is a category of AF utilized in this proposed system.

**Step 4:** Evaluate the next hidden layer's output using

$$Y_i = B_i + \sum C_i W_i \quad (16)$$

where $B_i$ denotes the bias value and $W_i$ specifies the weight between the input and the hidden layers.

**Step 5:** Here, the above three steps are carried out for each layer in the MDCNN. Finally, evaluate the output unit by adding up all the input signals' weights to obtain the output layer neurons' values:

$$R_i = B_i + \sum O_i W_j \quad (17)$$

where $O_i$ denotes the value of the layer that precedes the output layer, $W_j$ specifies the weights of the hidden layer, and $R_i$ indicates the output unit.

**Step 6:** This step compares the network output with the target value. The difference between these two values is called the error signal. This value is mathematically expressed as

$$E_r = D_i - R_i \quad (18)$$

where $E_r$ is the error signal and $D_i$ specifies the target output.

**Step 7:** Here, the value of the output unit is compared with the target value. The related error is determined. Based on this error, a value $\delta_i$ is computed, which is also utilized to transmit the error at the output back to all the other units in the network.

$$\delta_i = E_r[f(R_i)] \quad (19)$$

**Step 8:** The weight correction is done by employing the backpropagation algorithm. This relation is given as below:

$$W_{ci} = \alpha \delta_i (F_i) \quad (20)$$

where $Wc_i$ indicates the weight correction, $\alpha$ denotes the momentum, and $\delta_i$ is the error that is distributed in the network. The weight values are optimized utilizing the AEHO algorithm.

### D. Adaptive Elephant Herd Optimization

AEHO comprises the following assumptions:
- The population of elephants is split into clans. Every single clan contains a specific number of elephants.
- The male elephants normally leave their own clan and live alone.
- Each clan is guided by its eldest female elephant termed matriarch.

Matriarchs embrace the best solution in a herd of elephants, whereas the worst solution is decoded by the position of the male elephants. The elephant population is split into $j$ clans.

Each member $u$ of clan $c$ moves as per the supervision of matriarch $E_m$ with the highest fitness value in the generation. This process is given in equation (21).

$$P_{new,E_m,u} = P_{E_m,u} + \alpha(P_{best,E_m} - P_{E_m,v}) \times Rd \quad (21)$$

Here, $P_{new,E_m,u}$ represents the new position of $u$ in $c$, $P_{E_m,u}$ represents the old position, $P_{best,E_m}$ indicates the best solution of $E_m$, $\alpha \in [0,1]$ is the algorithm's parameter that ascertains the influence of the matriarch, and $Rd$ is a random number utilized to increase population diversity in the later stages of the algorithm. The best elephant's position in clan $P_{best,E_m}$ is updated by utilizing equation (22).

$$P_{new,E_m} = \beta \times P_{center,E_m} \quad (22)$$

Here, $\beta \in [0,1]$ denotes the second parameter of the algorithm, which controls the influence of $P_{center,E_m}$, as given by equation (23).

$$P_{center,E_m,d} = \frac{1}{u_{E_m}} \times \sum_{j=1}^{u_{E_m}} b_{E_m,j,d} \quad (23)$$

Here, $1 \leq d \leq D$ denotes the $d^{th}$ dimension, $D$ specifies the total space dimension, and $u_{E_m}$ indicates the number of elephants in clan $c$.

The male elephants that leave their clan are utilized for modeling exploration. In each clan $c$, some elephants with the worst fitness values are provided with new positions, as shown in equation (24):

$$P_{worst,E_m} = P(Pmin_{max} \times rand)_{min} \quad (24)$$

Where:

$P_{min}$ - Lower bound of the search space

$P_{max}$ - Upper bound of the search space

$rand \in [0,1]$ - A random number drawn from the uniform distribution

When the elephants' positions are evaluated, crossover and mutation operations are carried out to improve the optimization. Here, two-point crossover is employed. In this method, two points are selected on the parental chromosomes. The genes between these two points are interchanged between the parental chromosomes, and so the children's chromosomes are obtained. These points are evaluated as shown in equation (25) and (26).

$$x_1 = \frac{|P_{new,E_m}|}{3} \quad (25)$$

$$x_2 = x_1 + \frac{|P_{new,E_m}|}{2} \quad (26)$$

Then, the mutation is executed by swapping a number of genes from every chromosome with new genes. The swapped genes are the arbitrarily created genes with no repetition in the chromosome. This process is repeated until a solution with a better fitness value is obtained.

The pseudocode of the AEHO algorithm is given as below:

### E. Pseudocode for AEHO
*Input:* Weight values
*Output*: Optimized weight values
**Begin**
   **Initialization**
      Set generation counter $t = 1$,
      Set Maximum Generation $M_G$
      Population $X_i = \{X_1, X_2, \ldots X_n\}$
   **While** $t < M_G$ **do**
      Sort all the elephants according to their fitness
      **for** all clans $E_m$ in the population **do**
        **for** all elephants $j$ in the clan $E_m$ **do**
          Update $P_{E_m,n}$ and generate $P_{new,E_m,n}$ by using,
          $a_{new,e_m,n} = a_{e_m,n} + \alpha(a_{best,e_m} - a_{e_m,v}) \times r$
          **if** $P_{E_m,n} = P_{best,E_m}$ then
            Update $P_{E_m,n}$ and generate $P_{new,E_m,n}$ by
            using $P_{new,E_m} = \beta \times P_{center,E_m}$
          **end if**
        **end for**
      **end for**
      **for** all clans $E_m$ in the population **do**
        Perform crossover and mutation
        Replace the worst elephant in clan $e_m$ by using
        $P_{worst,E_m} = P(Pmin_{max} \times rand)_{min}$
      **end for**
      Evaluate population by the newly updated positions.
      $t = t + 1$
   **end while**
   **return** the best solution among all population
**End**

TABLE 1: SYMBOLS USED IN THE MODEL.

| Symbol | Abbreviations |
|---|---|
| $\delta$ | Function's initial value |
| $N_s$ | New solution |
| $R_e$ | Reflection |
| $V_s$ | Visibility |
| $Cr$ and $Br$ | Two chaos random numbers |
| $AV_B$ | Average value of the best points |
| $F_i$ | The input value |
| $A_f$ | The activation function |
| $C_i$ | Specifies the exponential of $F_i$ |
| $B_i$ | The bias value |
| $W_i$ | Specifies the weight between the input and the hidden layers |
| $O_i$ | The value of the layer that precedes the output layer |
| $W_j$ | Specifies the weights of the hidden layer |
| $R_i$ | Indicates the output unit. |
| $E_r$ | The error signal |
| $D_i$ | Specifies the target output |
| $W_{ci}$ | Indicates the weight correction |
| $P_{new,E_m,u}$ | Represents the new position of $u$ in $c$ |
| $u_{E_m}$ | Indicates the number of elephants in clan $c$. |
| $P_{min}$ | Lower bound of the search space |
| $P_{max}$ | Upper bound of the search space |
| $rand \in [0,1]$ | A random number drawn from the uniform distribution |

## IV. EXPERIMENTS AND RESULTS DISCUSSION
The simulation of the proposed IoT framework for heart disease prediction system is employed using the python, Android and Java platform.

### A. Simulation Setup
The proposed IoT framework has been porotype by integrating available hardware devices, microcontroller and LoRa communication hardware to transmit the data to cloud system. The patient's age and sex have been stored along with patient identification number in the system. The chest pain (cp) parameter is generated using pseudo number generated between 1-4 because of unavailability of device to measure the chest pain. The resting blood pressure(trestbps) parameter has been captured using Omron HeartGuid-bp8000m that provide data through cloud environment and alerts on the mobile phone [49]. The serum cholestoral and glucose level have been generated using pseudo number generated in the pre-determined range because of unavailability of wearable device. The electrocardiographic data has been measured using AD8232 heart monitor board [50]. The maximum heart rate achieved, oldpeak and slop is also captured from historical data of the patient. The data is capturing and processing is performed by Raspberry Pi single board computer [52]. The list of hardware for this experiment is described in table 2.

TABLE 2: HARDWARE USED IN THE MODEL.

| Hardware | Description |
|---|---|
| AD8232 | Analog Devices electrocardiographic board |
| HeartGuide BP8000m | Omron wearable smartwatch for blood pressure monitoring |
| Raspberry Pi-IV | 1.5GHz quad-core 64-bit ARM Cortex-A72 CPU |
| SX1272 | 900MHz LoRa transmitter and receivers |
| Client computer | Intel(R) Core™ i5-2400CPU @3.10 GHz PC |

### B. Dataset
The proposed system utilizes three dataset that includes Hungarian heart disease dataset, Framingham, and Public Health which is publicly available on the Internet [48]. This database comprises around 76 attributes, yet only a subset of 14 attributes was utilized in the published experiments as shown in table 3. The selected features are shown in table 5. In the prediction of heart disease, 13 attributes are used while the last attribute acts as the output or predicted attribute for the presence of heart disease in a person. The "#58num" field contain values from 0-4 for diagnosis of heart disease where the scaling refers to the severity of the disease (4 being the highest). The distribution of 'num' attribute among the 303 records is shown in Fig. 2.

TABLE 3: DESCRIPTION OF ATTRIBUTES FROM UCI DATASET

| Attribute | Description |
|---|---|
| **#3 (age)** | age in years |
| **#4 (sex)** | sex (1 = male; 0 = female) |
| **#9 (cp)** | chest pain type<br>1. typical angina<br>2. atypical angina<br>3. non-anginal pain<br>4. asymptomatic |
| **#10 (trestbps)** | resting blood pressure (in mm Hg on admission to the hospital) |
| **#12 (chol)** | serum cholestoral in mg/dl |
| **#16 (fbs)** | fasting blood sugar > 120 mg/dl) (1 = true; 0 = false) |
| **#19 (restecg)** | resting electrocardiographic results<br>0. normal<br>1. having ST-T wave abnormality (T wave inversions and/or ST elevation or depression of > 0.05 mV)<br>2. showing probable or definite left ventricular hypertrophy by Estes' criteria |
| **#32 (thalach)** | maximum heart rate achieved |
| **#38 (exang)** | exercise induced angina (1 = yes; 0 = no) |
| **#40 (oldpeak)** | ST depression induced by exercise relative to rest |
| **#41 (slope)** | the slope of the peak exercise ST segment<br>1. upsloping<br>2. flat<br>3. downsloping |
| **#44 (ca)** | number of major vessels (0-3) colored by flourosopy |
| **#51 (thal)** | 3 = normal; 6 = fixed defect; 7 = reversable defect |
| **#58 (num)** | diagnosis of heart disease (angiographic disease status)<br>0. absence (< 50% diameter narrowing)<br>1-4. Present of heart disease (> 50% diameter narrowing) |

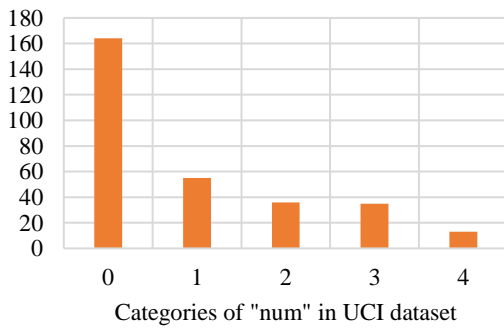

**FIGURE 2.** Distribution of UCI Dataset

The heart disease prediction system based on the MDCNN classifier is compared against the existing logistic regression (LR) and existing DLNN with respect to accuracy, error, precision, F₁ measure, Recall, specificity as show in table 9 and 10. The basic parameters that are evaluated are 'true positive' ($a_p$), 'true negative' ($a_n$), 'false positive' ($b_p$) and 'false negative' ($b_n$) values. The confusion matrix is shown as table 4.

TABLE 4: CONFUSION MATRIX

| Test Results | Truth | | Total |
|---|---|---|---|
| | Heart Disease | No Heart Disease | |
| Positive | $a_p$ | $b_p$ | $PPV = a_p/(a_p + b_p)$ |
| Negative | $a_n$ | $b_n$ | $NPV = b_n/(a_n + b_n)$ |

TABLE 5: SELECTED FEATURES FROM CLEVELAND DATASET

| Attribute | Description |
|---|---|
| #9 (cp) | chest pain type<br>1. typical angina<br>2. atypical angina<br>3. non-anginal pain<br>4. asymptomatic |
| #10 (trestbps) | resting blood pressure (in mm Hg on admission to the hospital) |
| #19 (restecg) | resting electrocardiographic results<br>1. normal<br>2. having ST-T wave abnormality (T wave inversions and/or ST elevation or depression of > 0.05 mV)<br>3. showing probable or definite left ventricular hypertrophy by Estes' criteria |
| #32 (thalach) | maximum heart rate achieved |
| #40 (oldpeak) | ST depression induced by exercise relative to rest |
| #41 (slope) | the slope of the peak exercise ST segment<br>1. upsloping<br>2. flat<br>3. downsloping |
| #58 (num) | diagnosis of heart disease (angiographic disease status)<br>0. absence (< 50% diameter narrowing)<br>1-4. Present of heart disease (> 50% diameter narrowing) |

The evaluation criteria for the simulation are shown as below:

*(a)* Accuracy (ACC): In general, accuracy relies on the way in which the data is gathered. It is judged by contrasting several measurements from the same or different sources. It is evaluated as equation (27).

$$accuracy = \frac{a_p + a_n}{a_p + a_n + b_p + b_n} \quad (27)$$

*(b)* Disease prevalence (DP): This is the chance of the disease being present in a person before health examination and is expressed as equation (28).

$$DP = \frac{a_p + b_n}{a_p + b_p + a_n + b_n} \quad (28)$$

*(c)* Precision, Positive predictive value (PPV): This is the probability of the patient with a positive screening test truly have the disease. The PPV can be evaluated as shown in equation (29).

$$PPV = \frac{a_p}{a_p + b_p} \quad (29)$$

*(d)* Negative predictive value (NPV): This indicates the probability of finding a patient with no heart disease risk and is evaluated as shown in equation (30).

$$NPV = \frac{a_n}{a_n + b_n} \quad (30)$$

*(e)* Sensitivity, Recall: This indicates the ability of finding a patient with heart disease risk and is evaluated as shown in equation (30).

$$sensivity = \frac{a_p}{a_p + b_n} \quad (31)$$

*(f)* Specificity: This can be computed by dividing true negative with total number of negative as show (32). The best specificity is defined by value 1.0 and worst by 0.0.

$$specificity = \frac{a_n}{a_n+b_p} \quad (32)$$

*(g)* $F_1$ Score: This can be expressed as shown in equation (33).

$$F_1 = 2 \cdot \frac{Precision \cdot Recall}{Precision + Recall} \quad (33)$$

The accuracy of the proposed MDCNN classifier is compared against the LR and DLNN for different number of records as shown in table 8. From table 8, it can be seen that the existing DLNN offers the worst performance. As shown in table 5, the selected features are varied on the basis of different feature selection algorithms. The performance is evaluated based on the number of records ranging from 303 to 4000. The analysis is shown graphically in Fig. 3.

TABLE 6: COMPARISON OF FEATURE SELECTION ALGORITHMS

| Feature selection algorithms | Selected Features |
|---|---|
| Relief | 13(1,2,3,4,5,6,7,8,9,10,11,12,13) |
| Info gain | 13(1,2,3,4,5,6,7,8,9,10,11,12,13) |
| Chi-squared | 13(1,2,3,4,5,6,7,8,9,10,11,12,13) |
| Filtered subset | 6(3,8,9,10,12,13) |
| One attribute based | 13(1,2,3,4,5,6,7,8,9,10,11,12,13) |
| Consistency based | 10(1,2,3,7,8,9,10,11,12,13) |
| Gain ratio | 13(1,2,3,4,5,6,7,8,9,10,11,12,13) |
| Filtered attribute | 13(1,2,3,4,5,6,7,8,9,10,11,12,13) |
| CFS | 8(3,7,8,9,10,11,12,13) |
| Genetic algorithm | 6(3,7,8,9,10,13) |
| MDCNN | 7(3,7,8,9,10,12,13) |

The performance of the proposed MDCNN is trained and evaluated with UCI dataset, Framingham, coronary heart disease (CHD), and Public Health Dataset. The proposed model selects the features with better accuracy.

TABLE 7: FEATURES SELECTED AND ACCURACY OF MDCNN

| Dataset | # Records | Total Features | Features selected | Accuracy |
|---|---|---|---|---|
| UCI | 303 | 16 | 7 | 93.3 |
| Framingham | 4000 | 16 | 7 | 98.2 |
| Public Health | 1025 | 14 | 8 | 97.6 |
| Sensor Data | 900 | 16 | 6 | 96.30 |

TABLE 8: CLASSIFIERS ACCURACY ANALYSIS

| Dataset | # Records | Classifiers | Accuracy |
|---|---|---|---|
| UCI | 303 | DLNN | 81.8 |
| | | LR | 87.8 |
| | | MDCNN | 93.3 |
| Framingham | 4000 | DLNN | 83.8 |
| | | LR | 88.3 |
| | | MDCNN | 98.2 |
| Public Health | 1025 | DLNN | 81.6 |
| | | LR | 84.6 |
| | | MDCNN | 97.6 |
| Sensor Data | 900 | DLNN | 82.4 |
| | | LR | 83.6 |
| | | MDCNN | 96.30 |

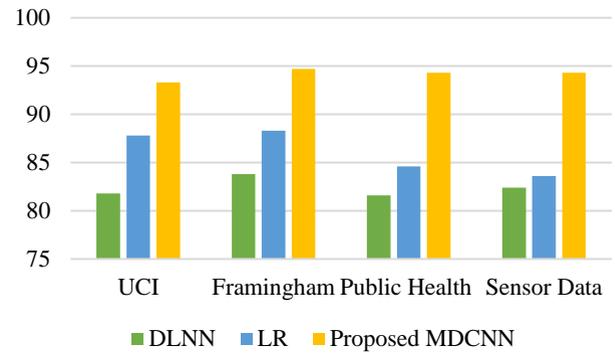

FIGURE 3. Accuracy of proposed algorithm

Fig. 3 shows that for maximum number of records, the MDCNN achieves 98.2 accuracy. In contrast, the existing LR and DLNN have lower accuracies of 88.3 and 81.6, respectively. Likewise, the proposed MDCNN achieves a higher performance for all the remaining 4000 records.

TABLE 9. PERFORMANCE ANALYSIS

| Models | ACC | Error | Precision | $F_1$ | Recall | Specificity |
|---|---|---|---|---|---|---|
| DLNN | 83.8 | 16.2 | 89.6 | 90.2 | 91.1 | 60.2 |
| LR | 88.3 | 11.7 | 90.1 | 90 | 92.8 | 74.2 |
| MDCNN | 98.2 | 1.8 | 95.1 | 95 | 97.8 | 92.6 |

TABLE 10: COMPARATIVE ANALYSIS

| Authors | Method | Accuracy |
|---|---|---|
| Senthilkumar et al. [53] | HRFLM | 88.7 |
| Xiao Liu et al. [43] | RFRS | 92.59 |
| M. Liu et al. [45] | LSTM | 97 |
| Proposed | MDCNN | 98.2 |

The work has been compared with state-of-art available that is close to this work as shown in table 10. The comparative analysis of the proposed MDCNN demonstrate the accuracy of 98.2 which is better than the existing methods. Table 11 present analysis on predictive values Vs disease prevalence.

TABLE 11: ANALYSIS OF PPV AND NPV

| DP | #Records | PPV (%) | | |
|---|---|---|---|---|
| | | DLNN | LR | MDCNN |
| 85 | 303 | 87.1 | 90.1 | 97.1 |
| 89 | 900 | 87.5 | 91.2 | 97.8 |
| 95 | 1025 | 88.1 | 92.1 | 98.1 |
| 98 | 4000 | 89.2 | 92.5 | 98.4 |

| DP | #Records | NPV (%) | | |
|---|---|---|---|---|
| | | DLNN | LR | MDCNN |
| 85 | 303 | 82.5 | 89.2 | 98.1 |
| 89 | 900 | 87.3 | 90.1 | 98.8 |
| 95 | 1025 | 88.1 | 92.3 | 99.2 |
| 98 | 4000 | 89.2 | 93.6 | 99.3 |

Table 11 compares the MDCNN performance with the existing LR and DLNN performance based on PPV and NPV. If a system has the best NPV and PPV, then the system is regarded as a good system. For example, for 4000 records, the proposed MDCNN has a PPV of 98.4, whereas the existing LR has a PPV of 92.5 and the existing DLNN has a PPV of 89.2, which are both lower than the PPV of the proposed method. The interpretation is that those who had a positive screening test, the probability of heart disease was 98.4. Next, for 4000 records, the MDCNN has an NPV of 99.3, whereas the existing LR and MDCNN have an NPV of 93.6 and 89.2 respectively, which are both lower than the NPV of the proposed MDCNN. This reveals that the proposed MDCNN has better performance than the existing methods. The table 11 also shows that higher the disease prevalence, the higher the PPV. When the prevalence of disease is low, the PPV will also be low.

## V. CONCLUSION AND FUTURE WORK

Wearable technologies can be utilized effectively in healthcare industry, particularly in chronic heart disease. The monitoring and prediction systems can help to save many lives by instant intervention specially when patient is located at remote place where medical facilities are not present.

The prediction of heart disease survivability is a challenging task. Previous prediction systems of heart attacks used certain techniques to diagnose heart disease. This paper proposed a wearable IoT enabled heart disease prediction system using the MDCNN classifier. Then, heart disease is predicted in the next three stages: (a) pre-processing, (b) feature selection, and (c) classification. The features are selected by using the MCFA, and normal and abnormal heart functioning is diagnosed by using the MDCNN. The performance of the proposed system was examined by utilizing the UCI dataset. The proposed MDCNN classifier also offers a higher level of accuracy than the existing approaches. The results demonstrate that the proposed methodology provides a higher level of accuracy than the other approaches. In the future, we will perform more experiments to increase the performance of these predictive classifiers for heart disease diagnosis by using others feature selection algorithms and optimization techniques. Also, the proposed work will be trained and tested with fully wearable devices which will be available in the market.

## ACKNOWLEDGMENT

The authors would like to thank University of Bisha, Bisha, Kingdom of Saudi Arabia for the facilities and support provided during the research.